\documentclass[journal,10pt]{IEEEtran}
\usepackage{diagbox}
\usepackage{graphicx}
\usepackage[cmex10]{amsmath}
\usepackage{cases}
\usepackage[tight,footnotesize]{subfigure}
\usepackage{amsthm} %proof
\usepackage{amsmath}
\usepackage{cite}
\usepackage{amssymb}
\usepackage{algorithm}
\usepackage{algorithmic}
\usepackage{xcolor}
\usepackage{stfloats}
\usepackage{multirow}
\usepackage{CJK}
\usepackage{subeqnarray}
\usepackage{longtable}
\usepackage{multicol}
\usepackage{extpfeil}
\usepackage{verbatim}
\allowdisplaybreaks
\newtheorem{theorem}{\bf{Theorem}}

\ifCLASSINFOpdf
\else
\fi
\begin{document}
\title{\huge{Edge-Assisted V2X Motion Planning and Power Control\\Under Channel Uncertainty}}

\author{Zongze Li, Shuai Wang, Shiyao Zhang, Miaowen Wen, Kejiang Ye,\\ Yik-Chung Wu,~\IEEEmembership{Senior Member,~IEEE}  and Derrick Wing Kwan Ng,~\IEEEmembership{Fellow,~IEEE} 
\thanks{
\scriptsize
Zongze Li is with Peng Cheng Laboratory, Shenzhen 518038, China (e-mail: lizz@pcl.ac.cn).
Shuai Wang and Kejiang Ye are with Shenzhen Institute of Advanced Technology, Chinese Academy of Sciences, Shenzhen 518055, China (e-mail: s.wang@siat.ac.cn; kj.ye@siat.ac.cn).
Shiyao Zhang is with the Academy for Advanced Interdisciplinary Studies, Southern University of Science and Technology, Shenzhen, China (e-mail: syzhang@ieee.org).
Miaowen Wen is with the School of Electronic and Information Engineering, South China University of Technology, Guangzhou 510640, China (e-mail: eemwwen@scut.edu.cn).
Yik-Chung Wu is with the Department of Electrical and Electronic Engineering, The University of Hong Kong, Hong Kong (e-mail: ycwu@eee.hku.hk).
Derrick~Wing~Kwan~Ng is with the School of Electrical Engineering and Telecommunications, the University of New South Wales, Australia (e-mail: w.k.ng@unsw.edu.au).
Corresponding Author: Shuai Wang. 
}
}


\maketitle

\begin{abstract}
Edge-assisted vehicle-to-everything (V2X) motion planning is an emerging paradigm to achieve safe and efficient autonomous driving, since it leverages the global position information shared among multiple vehicles.
However, due to the imperfect channel state information (CSI), the position information of vehicles may become outdated and inaccurate. 
Conventional methods ignoring the communication delays could severely jeopardize driving safety. To fill this gap, this paper proposes a robust V2X motion planning policy that adapts between competitive driving under a low communication delay and conservative driving under a high communication delay, and guarantees small communication delays at key waypoints via power control.
This is achieved by integrating the vehicle mobility and communication delay models and solving a joint design of motion planning and power control problem via the block coordinate descent framework. 
Simulation results show that the proposed driving policy achieves the smallest collision ratio compared with other benchmark policies. 
\end{abstract}



% \fi
\IEEEpeerreviewmaketitle

\section{Introduction}
Autonomous driving has attracted great attention from both academia and industry for its potential benefits on safety, accessibility, and efficiency. 
One fundamental task for autonomous driving is efficient and safe motion planning, i.e., to generate an optimal trajectory that not only attains the maximum system energy efficiency but also satisfies various safety constraints arising from traffic rules and operational limits of the vehicles \cite{J_Smith20Platoon}. Conventional motion planning strongly relies on perception techniques that generate real-time positions of surrounding obstacles. However, due to the sophisticated road environment and time-varying weather conditions, accurate positions cannot always be obtained via local sensing.

Vehicle-to-everything (V2X) motion planning has great potential to reduce the sensing uncertainty and to improve the safety for autonomous driving~\cite{FLCAV,baseline1,baseline2}. 
It consists of 4 main steps: 1) each vehicle obtains its ego position and velocity using onboard sensors; 2) all the vehicles share the information of their positions and velocities with an edge server (ES); 3) the ES fuses the information and forwards the fused information to all vehicles; 4) the ego-vehicle navigates itself to reach the goal while avoiding collisions with others. However, due to the inevitable wireless channel uncertainty~\cite{J_DJ08Selected}, V2X motion planning suffers from communication delays during the information sharing stage, and the position and velocity information may become outdated and inaccurate. As a result, the conventional method \cite{baseline1} ignoring communication uncertainties may lead to a dangerous driving policy. While the recent work in~\cite{baseline2} considering communication uncertainties could mitigate the above issue to a certain extent, a collision may still occur due to the potential V2X outage, especially at key waypoints, e.g., merging and lane-change.
Therefore, it is necessary to not only exploit, but also control the channel uncertainty for V2X motion planning, through the joint control of vehicle motion and transmit power, which is still an open problem.

In this paper, the impact of V2X physical layer designs on motion planning is concisely characterized for the first time. By integrating the communication delay and the vehicle mobility models, a stochastic programming is formulated to achieve a safe and efficient motion planning via the proposed safe distance and transmit power control scheme, which aims to allocate more power resources and extra safe distance margin at key turning points. The joint safety distance and power control problem is nontrivial due to the nonlinear couplings between motion dynamics and transmit power variables as well as the probabilistic function arising from the communication delay model. 
To address this challenging problem, this paper transforms the intractable probabilistic function into a deterministic one by leveraging the Marcum Q-function~\cite{Dig_Fading05}. 
Then, the resulting problem is tackled via the block coordinate descent (BCD) framework, where the interior-point method~\cite{B_Nesterov04Opt} and the projected-gradient (PG) method~\cite{B_Bertseka97NP} are employed to handle the corresponding subproblems, respectively. Finally, simulation results show that the proposed driving policy achieves the smallest collision ratio compared with other benchmark policies. 

The remainder of this paper is organized as follows. Section II introduces the system model and formulates the regularized MPC problem with channel uncertainty. Section III derives the solution to the optimization problem. Simulation results are presented in Section IV and the conclusion is drawn in Section V.

\section{System Model and Problem Formulation}

\begin{figure}[t!]
	\centering
	\includegraphics[scale=0.32]{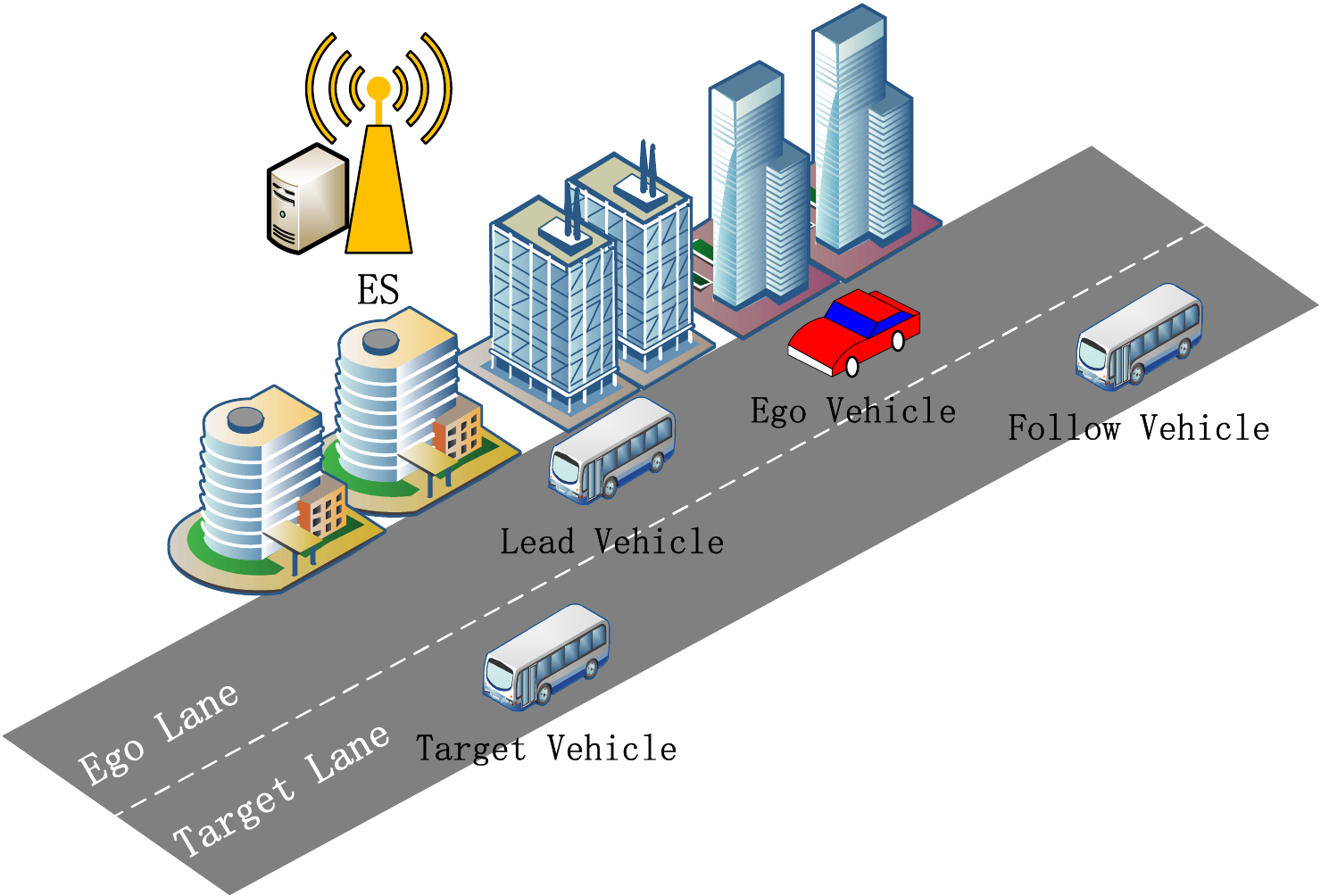}
	\caption{V2X motion planning for the lane-change scenario.}\label{fig:SystemModel}
\end{figure}

We consider a lane-change driving scenario with one edge server (ES), four vehicles, and two lanes, as shown in Fig.~\ref{fig:SystemModel}.
The ego vehicle (EV) is intended to take a lane-change motion to its left-hand side. During the movement, three surrounding vehicles are considered for motion planning, including the lead vehicle (LV) in the ego lane (EL), a target vehicle (TV) and a follow vehicle (FV) in the target lane (TL). All related parameters are disambiguated in Table~\ref{tab:PararTable}.

\vspace{-0.1in}
\subsection{Lane-Change Vehicle Mobility Model}

The EV adopts a model predictive control (MPC) policy to perform trajectory planning within a time horizon (e.g., $6\,$seconds), where the time horizon is divided into $K$ equal-length time slots with $\Delta t$ being the time interval between two consecutive slots.
At time instant $k$, the state of the EV is represented by $x_k,y_k,\theta_k$~\cite{J_Qian16Optinal_ITS}, where $x_k,y_k$ are the longitudinal and lateral coordinates, respectively, and $\theta_k\in[0,2\pi]$ is the yaw angle.
On the other hand, the control variables of the EV at time $k$ are the velocity $v_k$ and the yaw rotation rate $\omega_k$.
Considering the vehicle motion constraint, these state variables and control variables need to satisfy the vehicle dynamics~\cite{J_Qian16Optinal_ITS}
\begin{equation}\label{eq:linearize_cons}
\left\{\begin{array}{ll}
x_k-x_{k-1}=v_k \cos (\theta_k)\Delta t,\\
y_k-y_{k-1} = v_k\sin (\theta_k)\Delta t,\\
\theta_k-\theta_{k-1} = \omega_k\Delta t.
\end{array}\right.
\end{equation}

In addition, due to the physical limitations of the vehicle, the following boundary constraints are imposed on the control variables:
\begin{equation}\label{ineq:bounded_control}
\left\{\begin{array}{ll}
v_\mathrm{min} \leq v_k\leq v_\mathrm{max}, \\
\omega_\mathrm{min} \leq \omega_k \leq \omega_\mathrm{max},
\end{array}\right.
\end{equation} 
where $v_\mathrm{min}$, $v_\mathrm{max}$, $\omega_\mathrm{min}$ and $\omega_\mathrm{max}$ are the lower and upper bounds on the velocity and yaw rate for the EV, respectively.

Finally, the EV must avoid collision with the surrounding vehicles. 
Denote the longitudinal positions of the LV, the TV, and the FV by $x^\mathrm{LV}_k$, $x^\mathrm{TV}_k$, and $x^\mathrm{FV}_k$, respectively.
Then, the logical obstacle avoidance constraints are expressed by \cite{J_xi2020efficient}
\begin{equation}\label{cons:safe_distance}
\left\{\begin{array}{ll}
x^\mathrm{LV}_k-x_k\geq D, &~~\mathrm{if}~\mathrm{EV} \in \mathrm{EL},\\
x^\mathrm{TV}_k-x_k\geq D, &~~\mathrm{if}~\mathrm{EV} \in \mathrm{TL},\\
x_k-x^\mathrm{FV}_k\geq D, &~~\mathrm{if}~\mathrm{EV} \in \mathrm{TL},
\end{array}\right.
\end{equation} 
where $D$ is the minimum distance to avoid a collision.

Having specified the mobility constraints, the remaining step is to portray the goal of motion planning. 
By generating a sequence of EV positions $\{x_k,y_k\}_k^K$ that are close to the target waypoints $\{\hat{x}_k,\hat{y}_k\}_k^K$, the MPC 
policy aims to provide a safe and efficient driving policy by solving a finite horizon optimal control problem~\cite{C_sun2018fast}.
To facilitate subsequent optimization, we define a variable 
$\mathbf{u}_k:=[x_k-\hat{x}_k,y_k-\hat{y}_k]^T$ as the distance between the planned positions and the target positions and variable $\mathbf{c}_k:=[v_k-v_{k-1},\omega_k-\omega_{k-1}]^T$ as the change of the control parameters between two successive time slots. Since $\mathbf{u}_k$ is the distance incremental variable and $\mathbf{c}_k$ is the control variable, they measure the driving state and driving behaviour, respectively. Accordingly, the cost function to be minimized under the MPC policy is defined as~\cite{C_sun2018fast}
\begin{align}\label{eq:MPC_CF_lane}
\sum_{k=1}^{K}\left(\mathbf{u}_k^T\mathbf{W}_e\mathbf{u}_k
+\mathbf{c}_k^T\mathbf{W}_u\mathbf{c}_k\right),
\end{align}
where $\mathbf{W}_e\in \mathcal{R}^{2\times2}\succ \mathbf{0}$ and $\mathbf{W}_u\in \mathcal{R}^{2\times2}\succ \mathbf{0}$ are constant positive definite weight matrices.
It characterizes the total cost when performing the trajectory planning within a time horizon.

\begin{table}
	\centering
\caption{The parameters for the proposed MPC policy }\label{tab:PararTable}
\begin{tabular}{c|c}
	\hline
	Symbol &  Definition\\
	\hline
	$\Delta t$ & The time interval between two consecutive slots\\
	$x_k$ & Longitudinal coordinate at time slot $k$\\
	$y_k$ & Lateral coordinate at time slot $k$\\
	$\hat{x}_k$ & Target longitudinal coordinate at time slot $k$\\
	$\hat{y}_k$ & Target lateral coordinate at time slot $k$\\
	$\theta_k$ &  Yaw angle at time slot $k$\\
	$\omega_k$& Yaw rotation rate at time slot $k$\\
	$v_k$ & Velocity at time slot $k$\\
	$v_\mathrm{min}$ & The lower bound on the velocity \\
	$v_\mathrm{max}$ & The upper bound on the velocity \\
	$\omega_\mathrm{min} $& The lower bound on the yaw rate \\
	$\omega_\mathrm{max} $& The upper bound on the yaw rate \\
	$D$ & The two-second rule distance to avoid collision\\
 	$d_{i,k}$ & The distance from vehicle $i$ to the ES at time slot $k$\\
	$L_0$ & The reference path loss when the Tx-Rx distance is 1\,m\\
 	$\alpha$ & The path loss exponent, between 2 and 5 \\
	$m_d$ & The safe distance margin\\
	$h_{i, k}$ & The uplink channel from vehicle $i$ to the ES at time slot $k$ \\
	$P_{i,k}$ & The transmit power \\
	$P_{\rm{max}}$ & The instantaneous maximum power\\
	$n_{i,k}$ & The receiver noise \\
	$\tilde{e}_{i,k}$ & The feedback error\\
	$R$ & The minimum data rate requirement \\
	$C_{i,k}$ & The target channel capacity from node $i$ to the ES \\
	$\beta$ & The feedback accuracy \\
	$\sigma^2$ & The variance of receiver noise \\
	$\Xi$ & The regularized cost function \\
	$\rho_k$ & The penalty factor at time slot $k$\\
	\hline
\end{tabular}
\end{table}

\subsection{Communication with Channel Uncertainty}
The MPC policy requires the input of $\{x^\mathrm{LV}_k, x^\mathrm{TV}_k, x^\mathrm{FV}_k\}$ as observed from \eqref{cons:safe_distance}. 
This information is obtained from the ES and other vehicles. In particular, the surrounding vehicles first upload their position information to the ES in the uplink and the ES forwards the messages to the EV in the downlink.
Due to limited transmit power at the vehicles, there exists a non-negligible physical-layer outage probability, $p^\mathrm{out}_{i,k}$, for uplink communication from node $i$ to the ES at time $k$, where  $i\in\mathcal{I}:=\{\rm{LV}, \rm{TV}, \rm{FV}\}$.\footnote{It is assumed that the outage probability of downlink transmission is negligible compared with that of uplink transmission due to high transmit power at the ES.}
When the communication outage happens, the message needs to be re-transmitted, which leads to V2X communication delays, denoted by $\{\tau_{i,k}\}$.
According to \cite{aoi}, 
the communication delay $\tau_{i,k}$ is a monotonically increasing function of outage $p^\mathrm{out}_{i,k}$, and their relationship can be approximately written as 
$\tau_{i,k}=\tau_0 p^\mathrm{out}_{i,k}$, where $\tau_0$ is the transmission time for each round of position information uploading.
In such a case, the position information would be inaccurate, which is given by
\begin{equation}
\left\{\begin{array}{ll}
x^\mathrm{LV}_k=\hat{x}^\mathrm{LV}_k + \Delta^\mathrm{LV}_k,\\
x^\mathrm{TV}_k=\hat{x}^\mathrm{TV}_k + \Delta^\mathrm{TV}_k,\\
x^\mathrm{FV}_k=\hat{x}^\mathrm{FV}_k + \Delta^\mathrm{FV}_k,
\end{array}\right.
\end{equation} 
where $\hat{x}^\mathrm{LV}_k,\hat{x}^\mathrm{TV}_k,\hat{x}^\mathrm{FV}_k$ are the ground truth positions and $\Delta^\mathrm{LV}_k,\Delta^\mathrm{TV}_k,\Delta^\mathrm{FV}_k$ are position estimation errors. 
Specifically, the position error model is given by
\begin{align}
\Delta^i_k\sim \mathcal{U}(-v_k(\tau_0 p^\mathrm{out}_{i,k}+t_{\mathrm{comp}}),v_k(\tau_0 p^\mathrm{out}_{i,k}+t_{\mathrm{comp}})),
\end{align}
where $\mathcal{U}$ is the uniform distribution, 
$v_k$ is the speed of the EV, 
and $t_{\mathrm{comp}}$ is the computation delay (including reading/writing messages and convergence time of algorithms), respectively. The above equation estimates the additional traveling distance of vehicles due to outdated information. 

V2X systems adopt the frequency division multiplexing transmission scheme to avoid the interference between the signals from different vehicles.
The system supports a 10\,MHz or 20\,MHz bandwidth channel, which is sufficient for 4 vehicles (each allocated at least a 2.5\,MHz width channel) to exchange position information.
We consider a quasi-static block Rayleigh fading model.\footnote{The ES is located at the computation center which is far from the moving vehicles. Hence, there is no dominant line-of-sight signal propagation between the ES and vehicles. There are a large number of statistically independent reflected and scattered paths with random amplitudes.}
The uplink channel coefficient from vehicle $i$ to the ES at time slot $k$ is denoted by $h_{i, k}$ with $h_{i, k} \sim \mathcal{CN}(0,\mu^2_{i,k})$ and $i\in\mathcal{I}$. 
Note that $\mu^2_{i,k}=L_0d_{i,k}^{-\alpha}$, where $L_0,d_{i,k},\alpha$ are given in Table~\ref{tab:PararTable}.
The value of $\mu^2_{i,k}$ is predictable and obtained via the channel estimations as it represents large-scale fading and can be viewed as a constant during the lane change procedure.
The wireless transmission model is
\begin{equation} \label{eq:yb}
y_{i, k}=\sqrt{P_{i,k}}h_{i, k}s_{i, k}+n_{i, k},\quad i\in\mathcal{I},
\end{equation}
where $y_{i, k}$, $P_{i, k}$, $s_{i,k}$, and $n_{i,k}$ are the received signal, the transmit power, the transmitted symbol, and the receiver noise when node $i$ transmits information to its target receiver, respectively, with $\mathbb{E}\{|s_{i,k}|^2\}=1$ and $n_{i,k}\sim \mathcal{CN}(0,\sigma^2)$.

In practical V2X systems, the estimated CSIs $\{\hat{h}_{i, k}\}$ are obtained from feedbacks of target vehicles, who estimate the CSI by exploiting the channel reciprocity and the received training sequence. Due to the feedback errors, the CSIs at the ES are imperfect and perturbed by the feedback error $\tilde{e}_{i,k}$. In this case, given the estimated value $\hat{h}_{i,k}$, the channel gain $h_{i,k}$ is given by~\cite{J_Hyadi16}
\begin{equation}\label{eq:h_AB}
{h_{i,k}}=\mu_{i,k}\sqrt{\beta}\hat{h}_{i,k}+\mu_{i,k}\sqrt{1-\beta}\tilde{e}_{i,k},~i\in\mathcal{I},
\end{equation}
where $\tilde{e}_{i,k}\sim \mathcal{CN}\left(0,1\right)$ is the error at node $i$ and time slot $k$ and is independent of $h_{i,k}$. The parameter $\beta\in\left(0,1\right]$ represents the feedback accuracy and is obtained by calculating the correlation coefficient between $h_{i, k}$ and $\hat{h}_{i,k}$. 
The channel model reduces to the perfect CSI case when $\beta=1$.

From~\eqref{eq:yb}, the channel capacity from node $i$ to its target is $C_{i,k}=\log_2\left(1+P_{i,k}|h_{i,k}|^{2}/\sigma^2\right)$. 
To guarantee that the positions of all vehicles can be reliably received by the ES and that the decision of the server is successfully received by the EV, 
$C_{i,k}$ must satisfy $C_{i,k}\geq R$, where $R$ is the minimum data-rate requirement for sharing position information.
However, since $C_{i,k}$ is inaccurately known due to the imperfect knowledge of $\hat{h}_{i,k}$, a reliable communication link cannot always be guaranteed as it is possible that $R>C_{i,k}$. Hence, an appropriate measure is the outage probability, which is expressed as
\begin{equation}\label{eq:p_ro}
\begin{split}
p^\mathrm{out}_{i,k}(P_{i,k})&\!=
\mathrm{Pr}\left\{\left.C_{i,k}<R\right|\hat{h}_{i,k}\right\}\\
&\!=\mathrm{Pr}\left\{\left.\log_2\left(1+{P_{i,k}|h_{i,k}|^{2}}/{\sigma^2}\right)<R\right|\hat{h}_{i,k}\right\}.
\end{split}
\end{equation}

\subsection{Regularized MPC Problem Formulation}

Due to the V2X transmission delay, the perfect information of the surrounding vehicles may not be available and the safe following distance parameter $D$ in \eqref{cons:safe_distance} cannot guarantee collision avoidance during the lane change. To ensure a safe EV movement due to the wireless channel uncertainty, this paper proposes to add an extra safe distance margin $m_d$, which modifies \eqref{cons:safe_distance} into
\begin{equation}\label{cons:safe_distance_new}
\left\{\begin{array}{ll}
x^\mathrm{LV}_k-x_k\geq D+m_d, &~~\mathrm{if}~\mathrm{EV} \in \mathrm{EL},\\
x^\mathrm{TV}_k-x_k\geq D+m_d, &~~\mathrm{if}~\mathrm{EV} \in \mathrm{TL},\\
x_k-x^\mathrm{FV}_k\geq D+m_d, &~~\mathrm{if}~\mathrm{EV} \in \mathrm{TL}.
\end{array}\right.
\end{equation} 
Different from parameter $D$, $m_d$ is an optimization variable that should be optimized based on the channel state information and vehicle dynamics. The optimized $m_d$ should satisfy the following properties: 1) If $p^\mathrm{out}_{i,k}$ is large, then $m_d$ should also be large; 2) If $p^\mathrm{out}_{i,k}=0$ for all $(i,k)$, then $m_d=0$; 3) The benefit of increasing $m_d$ is reduced as $m_d$ increases. To design an MPC considering the communication outage, we propose a regularized cost function $\Xi(m_d, P_{i,k})$ to (4) as follows:
\begin{align}
&\Xi(m_d, P_{i,k})=
\frac{ p^\mathrm{out}_{i,k}(P_{i,k})}{1-e^{-m_d}},
\end{align}
which has the following properties: 1) $\Xi$ approaches infinity when  $m_d\rightarrow 0$, which encourages a positive $m_d$ to guarantee the safety; 2) $\Xi$ is monotonically increasing in $P_{i,k}$, which motivates the EV to select a small $m_d$ when transmit power $P_{i,k}$ is small; 3) $\Xi$ decreases slower as $m_d$ increases, which prevents the EV from choosing an exceedingly large $m_d$.

Based on the above discussions, our joint motion and power control optimization problem is formulated as 
\begin{subequations}\label{eq:Secrecy_Rate_Max}
	\begin{align}
	\mathcal{P}0:\min_{\mathcal{F}} &\sum_{k=1}^{K}\left(\mathbf{u}_k^T\mathbf{W}_e\mathbf{u}_k +
	\mathbf{c}_k^T\mathbf{W}_u\mathbf{c}_k
	+\rho_k\sum_{i\in\mathcal{I}}\frac{ p^\mathrm{out}_{i,k}(P_{i,k})}{1-e^{-m_d}}\right)\label{eq:max_SR} \\
	\mathrm{s.t.}~
	&\eqref{eq:linearize_cons},~\eqref{ineq:bounded_control},~\eqref{cons:safe_distance_new},~m_d\geq 0,\\
	&\sum_{k=1}^{K}P_{i,k}\leq P_\mathrm{max}, ~P_{i,k}\geq 0,~\forall i,k, \label{9c}
	\end{align}
\end{subequations}
where $\mathcal{F}=\{\{\mathbf{u}_k,\mathbf{c}_k,\{P_{i,k}\}_{i\in\mathcal{I}}\}_{k=1}^K,m_d\}$ and $\{\rho_k>0\}_{k=1}^K$ is the penalty factors. \eqref{9c} is the total power budget constraint with $P_{\rm{max}}$ being the power budget.\footnote{The instantaneous power constraint is not considered as vehicles are equipped with large RF ends and antennas.}

\emph{Remark 1}: The motion planning of vehicles and the power control of communication in $\mathcal{P}0$ are executed in the same time slot. This is because a common motion control frequency ranges from 20\,Hz to 60\,Hz \cite{editor} and the power control frequency is equal to the inverse of a channel coherence time slot, which ranges typically from 15\,ms to 50\,ms, also corresponding to a frequency of 20\,Hz to 60\,Hz.

\section{Optimization Solution to $\mathcal{P}0$}

\subsection{Outage Probability Transformation}
According to~\eqref{eq:h_AB}, when $\hat{h}_{i,k}$ is obtained, ${h_{i,k}}/{\mu_{i,k}}$ can be regarded as a complex Gaussian variable with mean $\sqrt{\beta}\hat{h}_{i,k}$ and variance $(1-\beta)$. Therefore, $|{h_{i,k}}/{\mu_{i,k}}|^{2}$ given $\hat{h}_{i,k}$ follows a non-central $\chi^2$ distribution with two degrees of freedom and the conditional cumulative distribution function (CDF) is
\begin{align} \label{eq:pdf_gamma_b}
F_{|{h_{i,k}}/\mu_{i,k}|^{2}|\hat{h}_{i,k}}\left(x\right)
=1-Q_1\left(\frac{\sqrt{\beta|\hat{h}_{i,k}|^{2}}}{\zeta},\frac{\sqrt{x}}{\zeta}\right),
\end{align} 
where $\zeta=\sqrt{(1-\beta)/2}$ and $Q_1(x,y)$ is the first-order Marcum Q-function~\cite[eq. 4.33]{Dig_Fading05}. 
Based on~\eqref{eq:pdf_gamma_b}, $p^\mathrm{out}_{i,k}(P_{i,k})$ in~\eqref{eq:p_ro} is given by
\begin{equation}\label{eq:Outage_ini_Pik}
p^\mathrm{out}_{i,k}(P_{i,k})
=F_{|{h_{i,k}}/\mu_{i,k}|^{2}|\hat{h}_{i,k}}\left(\frac{2^{R}-1}{P_{i,k}\mu^2_{i,k}}\sigma^2\right).
\end{equation}
Then, putting~\eqref{eq:Outage_ini_Pik} into $\mathcal{P}0$ yields
\begin{subequations}   
	\begin{align}
	\mathcal{P}1:\min_{\mathcal{F}}~&
\sum_{k=1}^{K}\mathbf{u}_k^T\mathbf{W}_e\mathbf{u}_k +\mathbf{c}_k^T\mathbf{W}_u\mathbf{c}_k\nonumber \\
&+\sum_{k=1}^{K}\sum_{i\in\mathcal{I}}\rho_k \frac{F_{|h_{i,k}/\mu_{i,k}|^{2}|\hat{h}_{i,k}}\left(\frac{2^{R}-1}{P_{i,k}\mu^{2}_{i,k}}\sigma^2\right)}{1-e^{-m_d}}  \\
\mathrm{s.t.}\quad
&\eqref{eq:linearize_cons},~\eqref{ineq:bounded_control},~\eqref{cons:safe_distance},~\eqref{9c},~m_d\geq 0.
	\end{align}
\end{subequations}
Since the constraint set of $\mathcal{P}1$ is decoupled when either $\{P_{i,k}\}$ or $\{\{\mathbf{u}_k,\mathbf{c}_k\}_{k=1}^K,m_d\}$ is fixed, the optimization problem $\mathcal{P}0$ can be decomposed into two subproblems under the BCD framework~\cite{J_yang20BSCA}. By solving problem $\mathcal{P}1$ at each time slot iteratively, the optimized control actions and power allocation can be directly obtained for motion planning, safe distance margin, and power control.

\subsection{Solving Problem $\mathcal{P}1$ under BCD Framework}

When $\{P_{i,k}\}$ are fixed, the subproblem of $\mathcal{P}1$ for optimizing efficient trajectory is given by
\begin{subequations}\label{obj:Determini_opt}
\begin{align}
	\mathcal{D}1:\!\min_{m_d,\{\mathbf{u}_k,\mathbf{c}_k\}_{k=1}^K}&
	\sum_{k=1}^{K}\left(\mathbf{u}_k^T\mathbf{W}_e\mathbf{u}_k \!+\!\mathbf{c}_k^T\mathbf{W}_u\mathbf{c}_k\!+\!\frac{\eta}{1-e^{-m_d}}\right)\\
	\mathrm{s.t.}\quad
	&\eqref{eq:linearize_cons},~\eqref{ineq:bounded_control},~\eqref{cons:safe_distance},~m_d\geq 0,
\end{align}
\end{subequations}
where $\eta = \sum_{i\in\mathcal{I}}\rho_k F_{|h_{i,k}/\mu_{i,k}|^{2}|\hat{h}_{i,k}}\left(\frac{2^{R}-1}{P_{i,k}\mu^{2}_{i,k}}\sigma^2\right)$. Since constraints~\eqref{eq:linearize_cons},~\eqref{ineq:bounded_control}, and~\eqref{cons:safe_distance} are affine constraints, together with $m_d\geq 0$,  the feasible set of $\mathcal{D}1$ is convex. On the other hand, notice that $\mathbf{W}_e$ and $\mathbf{W}_u$ are positive definite,  $\mathbf{u}_k\mathbf{W}_e\mathbf{u}_k^T$ and $\mathbf{c}_k\mathbf{W}_u\mathbf{c}_k^T$ are convex on $\mathbf{u}_k$ and $\mathbf{c}_k$, respectively. Therefore, $\sum_{k=1}^{K}(\mathbf{u}_k\mathbf{W}_e\mathbf{u}_k^T+ \mathbf{c}_k\mathbf{W}_u\mathbf{c}_k^T)$ is convex, since the sum of convex functions preserves convexity~\cite{Cov_Opt90}. 
Moreover, since $\eta $ is a constant value, $\sum_{k=1}^{K} \frac{\eta}{1-e^{-m_d}}$ is convex on $(0,+\infty)$~\cite{Cov_Opt90}. 
Therefore, the objective function of~\eqref{obj:Determini_opt} is convex and $\mathcal{D}1$ is a convex problem that can be efficiently solved via the interior-point method~\cite{B_Nesterov04Opt}.

On the other hand, when $\{\{\mathbf{u}_k,\mathbf{c}_k\}_{k=1}^K,m_d\}$ are fixed, the subproblem for updating $\{\{P_{i,k}\}_{i\in\mathcal{I}}\}_{k=1}^K$ is given by
\begin{subequations}
\begin{align}
	\mathcal{Q}1:\!\min_{\{\{P_{i,k}\}_{i\in\mathcal{I}}\}_{k=1}^K}&
\sum_{k=1}^{K}\sum_{i\in\mathcal{I}}\rho_k \frac{F_{|h_{i,k}/\mu_{i,k}|^{2}|\hat{h}_{i,k}}\left(\frac{2^{R}-1}{P_{i,k}\mu^{2}_{i,k}}\sigma^2\right)}{1-e^{-m_d}}\\
	\mathrm{s.t.}\quad &
	{\sum_{k=1}^{K}P_{i,k}\leq P_\mathrm{max},P_{i,k}\geq 0,\forall k,i.}
\end{align}
\end{subequations}
Since $k$ and $i$ are independent, the summation over $k$ and $i$ in $\mathcal{Q}1$ are interchangeable. Moreover, since the cardinality of set $\mathcal{I}$ is three and optimization variables  $\{P_{i,k}\}_{i\in\mathcal{I}}$ are independent of one another, subproblem $\mathcal{Q}1$ can be further reduced to three parallel subproblems with the $i^\mathrm{th}$ subproblem being
\begin{subequations}
\begin{align}
\mathcal{Q}1^{[i]}:~\min_{\{P_{i,k}\}_{k=1}^K}&
\sum_{k=1}^K\underbrace{\rho_k \frac{F_{|h_{i,k}/\mu_{i,k}|^{2}|\hat{h}_{i,k}}\left(\frac{2^{R}-1}{P_{i,k}\mu^{2}_{i,k}}\sigma^2\right)}{1-e^{-m_d}}}_{:= f\left(P_{i,k}\right)} \label{obj:Mar_Power_Q1}\\
\mathrm{s.t.}~~&
{\sum_{k=1}^{K}P_{i,k}\leq P_\mathrm{max},~P_{i,k}\geq 0,~\forall k} \label{eq:Q1_transmission}.
\end{align}
\end{subequations}
Since $\mathcal{Q}1^{[i]}$ has a continuously differentiable objective function and a linear feasible set, it can be solved by the projected-gradient (PG) method~\cite{B_Bertseka97NP}, which alternatively performs an unconstrained gradient descent step and computes the projection of the unconstrained update onto the feasible set of the optimization problem.
To be specific, the update of $\{P_{i,k}\}_{k=1}^{K}$ at the $l^\mathrm{th}$ iteration is given by
\begin{align}\label{eq:proj_grad}
&P_{i,k}\left(l+\frac{1}{2}\right)
=P_{i,k}(l)-\mathcal{A}^{(l)}\nabla_{P_{i,k}} f, ~\forall k,
\end{align}
where $\mathcal{A}^{(l)}$ is a backtracking step size chosen by the Armijo rule to guarantee the convergence~\cite[Prop. 2.3.3]{B_Bertseka97NP}, and $\nabla_{P_{i,k}} f$ is given by
\begin{equation}
\nabla_{P_{i,k}} f\!=\!\frac{-\rho_k(2^R-1)\sigma^2}{2(1-e^{-m_d})\zeta^2P_{i,k}^2}\frac{I_0\left(\sqrt{\frac{(2^R-1)\beta|\hat{h}_{i,k}|^{2}\sigma^2}{\mu^{2}_{i,k}P_{i,k}\zeta^{4}}}\right)}{\exp\left(\frac{\beta|\hat{h}_{i,k}|^{2}+\frac{2^{R}-1}{P_{i,k}\mu^{2}_{i,k}}\sigma^2}{2\zeta^2}\right)},
\end{equation}
where $I_0(x)$ is the zero-th order modified Bessel function of the first kind~\cite[eq. 8.445]{Table_Mathe00}.
On the other hand, to project $P_{i,k}\left(l+{1}/{2}\right)$ onto the feasible set of problem $\mathcal{Q}1^{[i]}$ to find its nearest feasible point $P_{i,k}(l+1)$, we have an equivalent optimization problem expressed as
\begin{align}\label{opt:proj_Q1}
&P_{i,k}(l+1) 
\!=\!\mathop{\arg\min}_{\{P_{i,k}\}_{k=1}^{K}\in \mathcal{S}_{1}}\! \sum_{k=1}^{K}\left\|P_{i,k}\!-P_{i,k}\left(l+\frac{1}{2}\right)\right\|^2,
\end{align}
where {$\mathcal{S}_{1}=\{\{P_{i,k}\}_{k=1}^{K}| \sum_{k=1}^{K}P_{i,k}\leq P_\mathrm{max},P_{i,k}\geq 0,\forall k\}$} is the domain of $\mathcal{Q}1^{[i]}$.
Since~\eqref{opt:proj_Q1} is a convex optimization problem, a closed-form solution can be derived based on Karush-Kuhn-Tucker (KKT) condition and is given by the following theorem, which can be proved by following a similar approach in~\cite[Proposition 2]{J_20NOMASecureGrad}.
\begin{theorem}\label{lem:Proj_gra_theta_k}
The optimal solution of~\eqref{opt:proj_Q1} is given by
	\begin{equation}\label{eq:pro_gra_optTheta}
\begin{aligned}
{P_{i,k}}^{\dagger}=\left[P_{i,k}(l+1/2)-\frac{\zeta^{\dagger}}{2}\right]^+,~~\forall k,
\end{aligned}
\end{equation}
where $\zeta^{\dagger}$ satisfies
$\sum_{k=1}^{K}\left[{P_{i,k}}(l+1/2)-{\zeta^{\dagger}}/{2}\right]^+=P_{\rm{max}}$ and can be obtained with the bisection search~\cite{B_Bertseka97NP}.
\end{theorem}
\noindent Based on~\eqref{eq:proj_grad} and Theorem~\ref{lem:Proj_gra_theta_k}, we can iteratively update $\{P_{i,k}\}_{k=1}^{K}$, where the convergence is guaranteed to be a stationary point of $\mathcal{Q}1^{[i]}$~\cite[Proposition 2.3]{B_Bertseka97NP}. Moreover, since $\mathcal{Q}1$ consists of three parallel subproblems $\mathcal{Q}1^{[i]}$ for each variables $\{P_{i,k}\}_{k=1}^{K},\forall i$, it can be solved in a parallel manner.

To sum up, with the feasible initial points in set $\mathcal{F}$, $\mathcal{P}1$ can be iteratively solved, where the proposed algorithm is summarized in Algorithm~\ref{Overall_Algo:Sove_P1}, and the convergence of the objective value of $\mathcal{P}1$ can be proved using the approach in \cite{J_yang20BSCA}. As for the computational complexity of Algorithm~\ref{Overall_Algo:Sove_P1}, it is dominated by the inner iteration in steps 3 and 4. For step 3, $\mathcal{D}1$ is solved via the interior-point method with the complexity order of $\mathcal{O}((2K)^{3.5})$. For step 4, since subproblem $\mathcal{Q}1^{[i]}$ is solved via the PG method, which only involves the first-order differentiation, it has $\mathcal{O}(K/\varrho)$ complexity order with an accuracy of $\varrho$~\cite{B_Bertseka97NP}. Since updating $\{\{P_{i,k}\}_{i\in\mathcal{I}}\}_{k=1}^K$ involves solving three parallel subproblems, the complexity of solving $\mathcal{Q}1$ is $\mathcal{O}(3K/\varrho)$. 
Together with the outer iterations, the complexity order of Algorithm~\ref{Overall_Algo:Sove_P1} is $\mathcal{O}(\mathcal{M}((2K)^{3.5}+3K/\varrho))$, where $\mathcal{M}$ is the outer iteration number to converge. The convergence behaviour of Algorithm~\ref{Overall_Algo:Sove_P1} is shown in Fig.~\ref{fig:perform_convergence}, where the proposed algorithm converges rapidly within 2 iterations under different vehicle speeds. 
The convergence time is around 1~s with MATLAB R2021b and can be reduced to tens of milliseconds by leveraging the more efficient programming languages (Python or C) in real-time operating systems.
\begin{algorithm}[t]
	\caption{Overall Algorithm for Solving $\mathcal{P}1$} 
	\begin{algorithmic}[1]\label{Overall_Algo:Sove_P1}		
		\STATE Initialize feasible points of $\{\{\mathbf{u}_k,\mathbf{c}_k,\{P_{i,k}\}_{i\in\mathcal{I}}\}_{k=1}^K,m_d\}$.
		\REPEAT
		\STATE Update $\{\{\mathbf{u}_k,\mathbf{c}_k\}_{k=1}^K,m_d\}$ for solving $\mathcal{D}1$ via the interior-point method. 
		\STATE Update $\{\{P_{i,k}\}_{i\in\mathcal{I}}\}_{k=1}^K$ by solving $\mathcal{Q}1^{[i]}$ in a parallel manner for all $i$ via the PG method.
		\UNTIL Stopping criterion is satisfied.	
	\end{algorithmic}
\end{algorithm}

\vspace{-0,1in}
\section{Simulation Results and Discussions}
In this section, we evaluate the driving performance of the proposed policy with channel uncertainty through simulations and experimental results. The width of each lane is 3.72 m based on the US standard. 
The starting locations of EV, LV, TV, FV are (20\,m, 1.85\,m), (30\,m, 1.85\,m), (40\,m, 5.55\,m), (13\,m, 5.55\,m), respectively.
The starting velocities of EV, LV, TV, FV are 7.2\,km/h, 5\,km/h, 25\,km/h, 7.9\,km/h, respectively.
The bandwidth is set to be 10 MHz with noise power spectral density being $\sigma^2=-96\,\mathrm{dBm/Hz}$ \cite{LTE_carrierFQ13}. 
The required spectral efficiency is set to $R=2\,\mathrm{bps/Hz}$, the maximum transmit power is $P_\mathrm{max}= 30\,\mathrm{dBm}$, and the large-scale fading parameter is set to $\mu^2_{i,k}=3.5$. The safe distance is $D=8.7\,$m according to the two-second rule (with a vehicle length of $4.7$\,m and a speed of 7.2\,km/h), $\Delta t=1\,$s, and the initial yaw angles for the LV, the TV, and the FV are all $0^{\mathrm{o}}$. According to 3GPP~\cite{LTE_carrierFQ13}, the transmission time is set to $\tau_0=50$ ms. The computation delay is $t_{\mathrm{comp}}=10$\,ms. We employ the Car Learning to Act (CARLA) simulation platform for high-fidelity environments and dynamics generation~\cite{FLCAV}. Each vehicle is equipped with a positioning sensor and a communication device. The penalty factor $\{\rho_k\}_{k=1}^6$ are set to $\rho_1=1$ and $\{\rho_k=10\}_{k=2}^6$. As such, the EV would be more conservative during its turning stage of the lane-change task. 

\begin{figure*}	
	\centering
	\subfigure[Convergence behaviour of Algorithm 1 with different velocities.]
{\label{fig:perform_convergence} %% label for second subfigure
\includegraphics[height=1.2in]{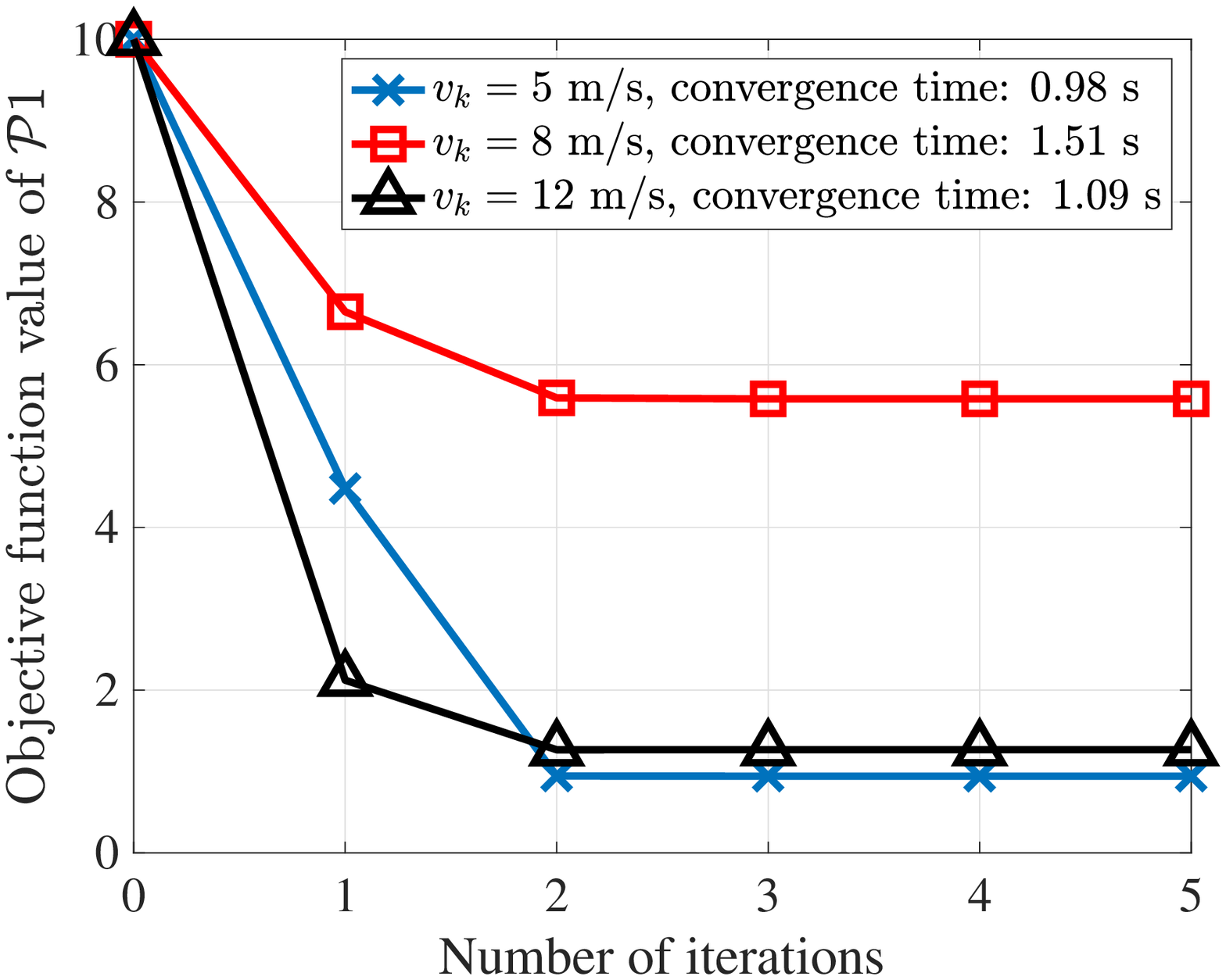}
}
\hspace{0.0in} 
\subfigure[Positions of all vehicles during lane-change movements.]{ 
		\label{fig:Trajectory_ego}%% label for first subfigure
		\includegraphics[height=1.2in]{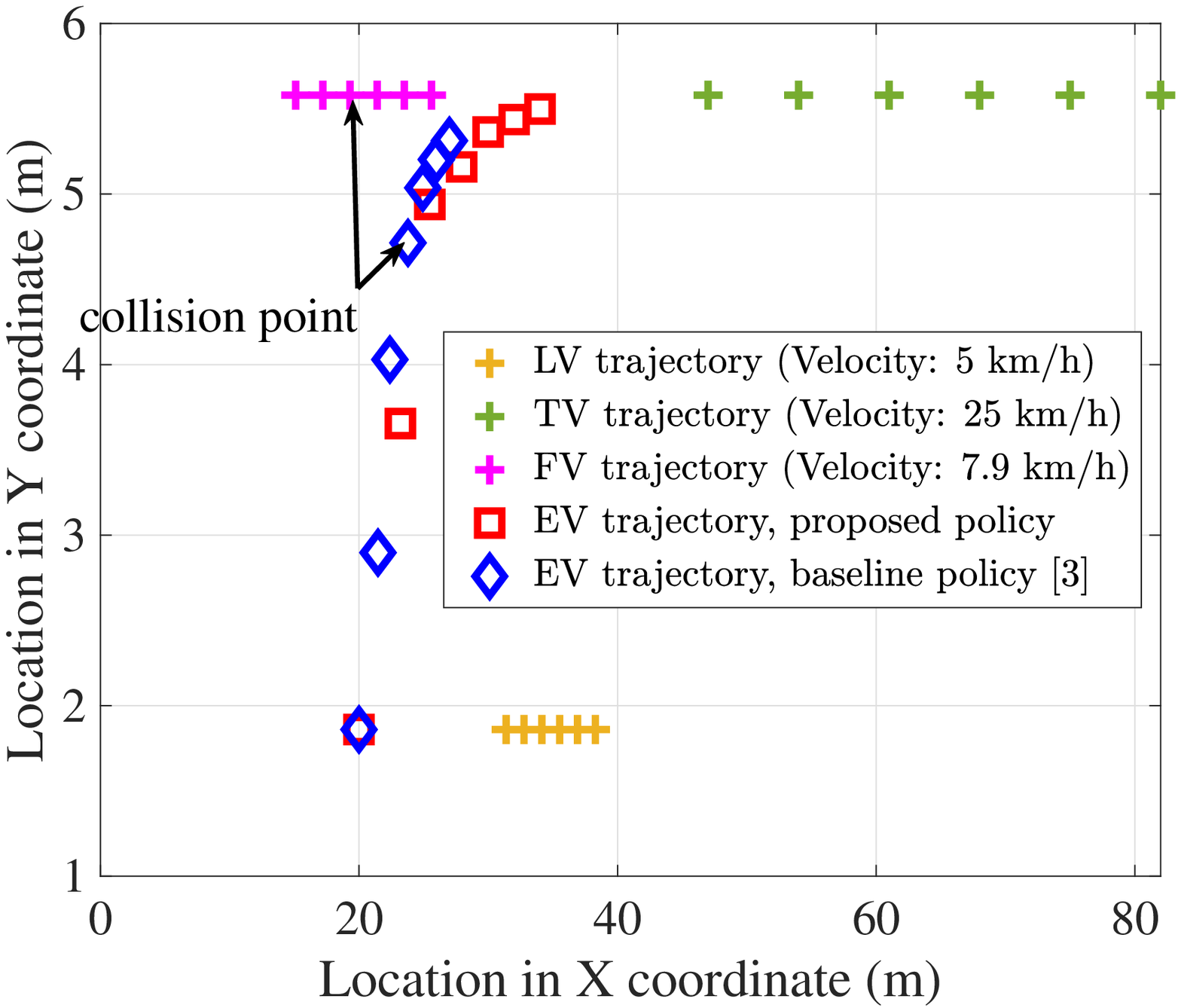}}
	\hspace{0.0in} 
	\subfigure[Positions of all vehicles during lane-change movements with high-speed surrounding vehicles. ]{
		\label{fig:Solution_30km} %% label for second subfigure
		\includegraphics[height=1.2in]{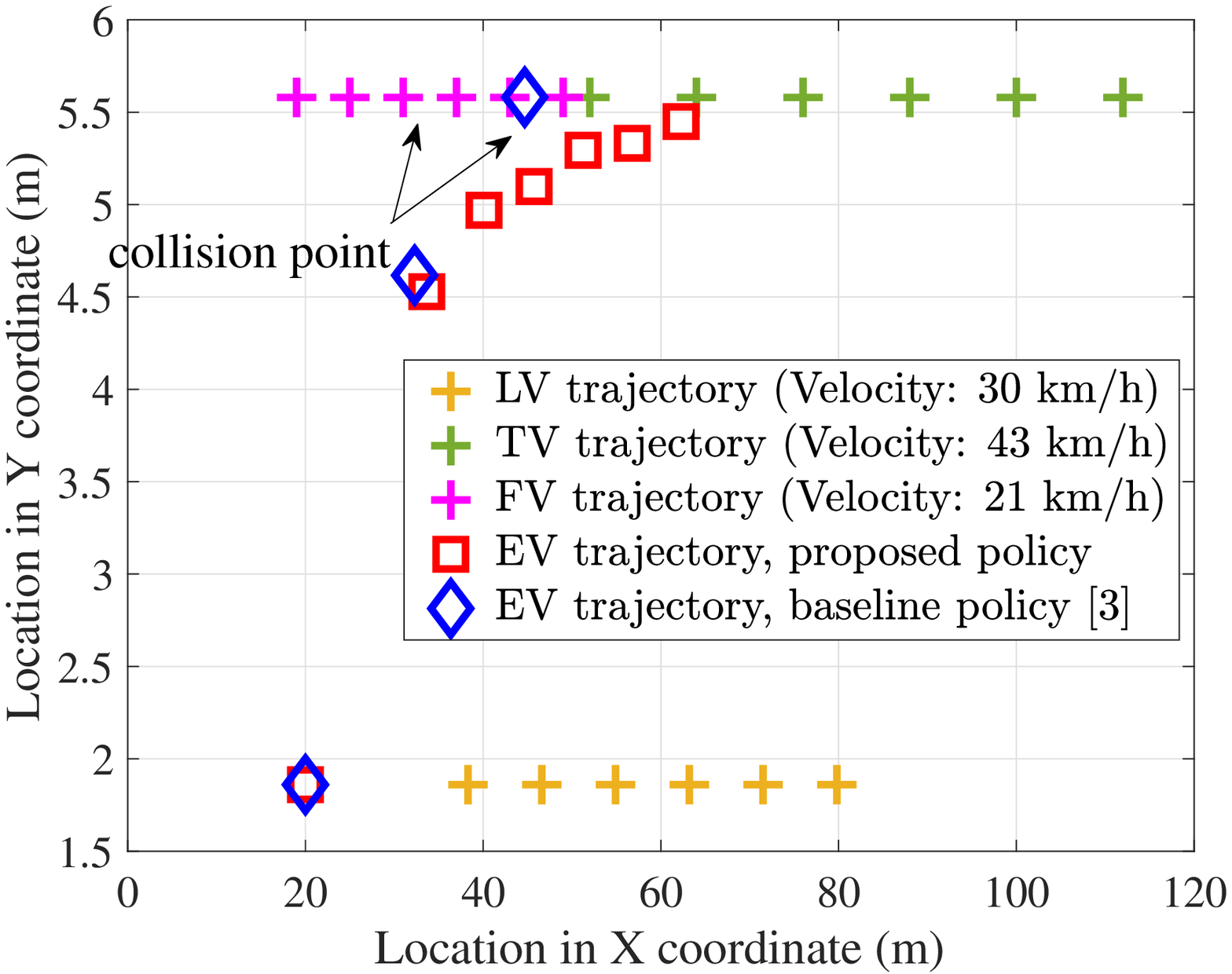}}
 \hspace{0.0in} 
 \subfigure[Positions of all vehicles during lane-change movements with variable-speed surrounding vehicles.]{
	\label{fig:Solution_accele_good} %% label for second subfigure
		\includegraphics[height=1.2in]{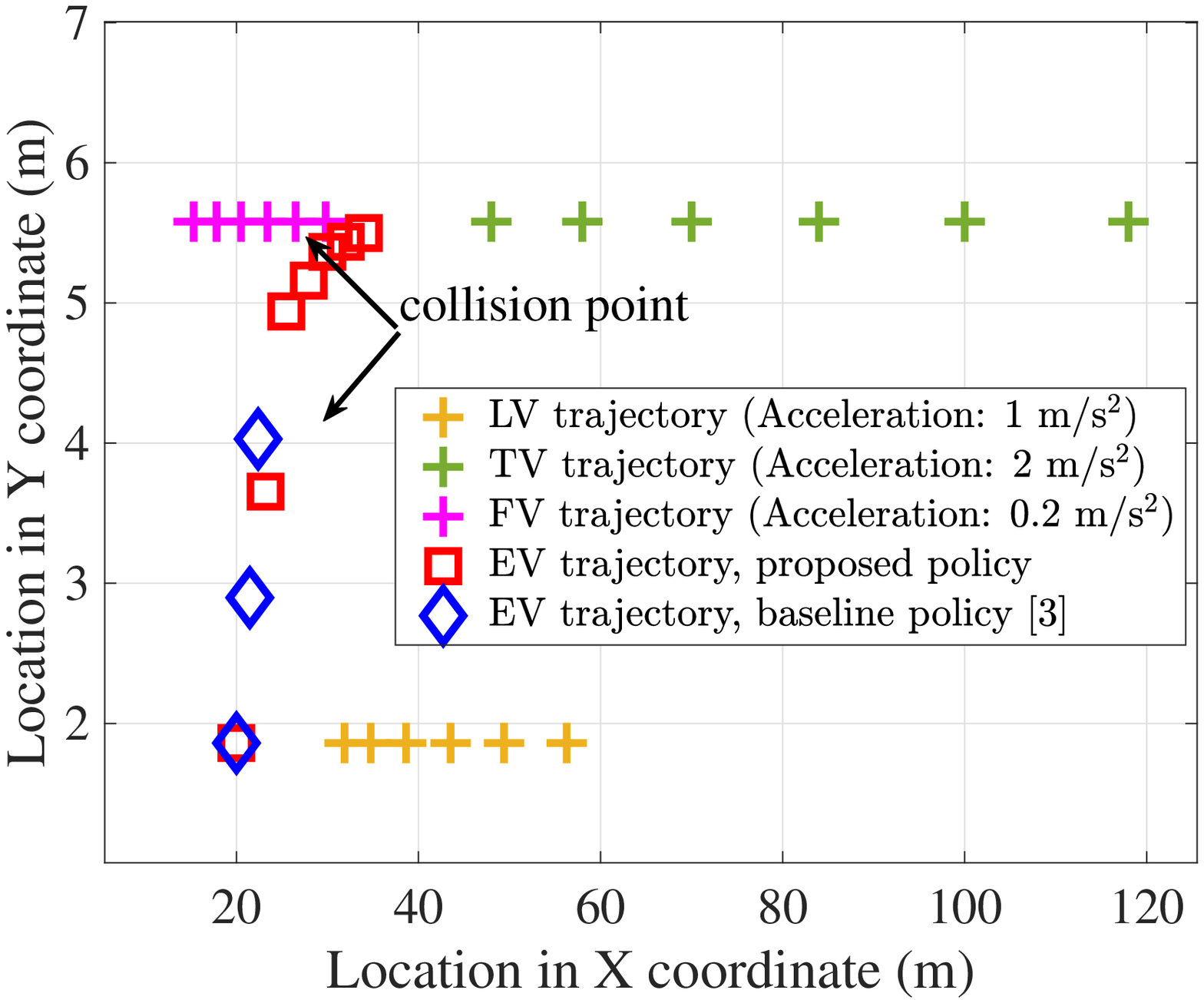}}  
		\hspace{0.0in}
	\caption{Convergence behaviour of the proposed algorithm and trajectories of all vehicles during lane-change with $\beta=0.3$ and $a=1$.} \label{fig:demo_LaneChange}
	\vspace{-0.1in}
\end{figure*}

\begin{figure*}	
	\centering
 		\subfigure[The relative velocities between the EV and its surrounding vehicles.]{\label{fig:Velocity_Rela}
	\includegraphics[height=1.2in]{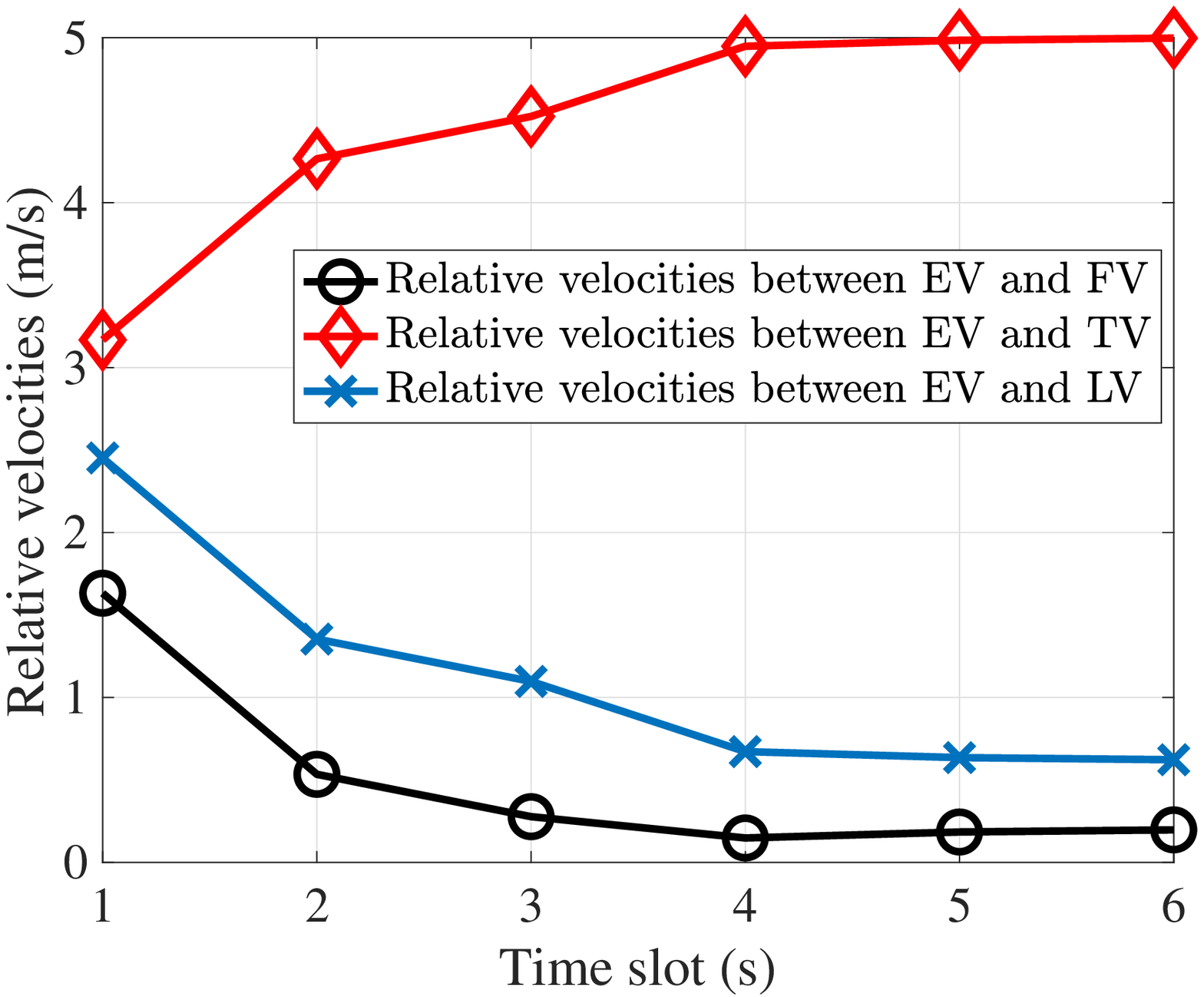}}
	\hspace{0.0in} 
		 \subfigure[Bird's-eye view of motion planning for the proposed and baseline policies~\cite{baseline1}.]{
		\label{fig:Collis_outage} %% label for second subfigure
		\includegraphics[height=1.1in]{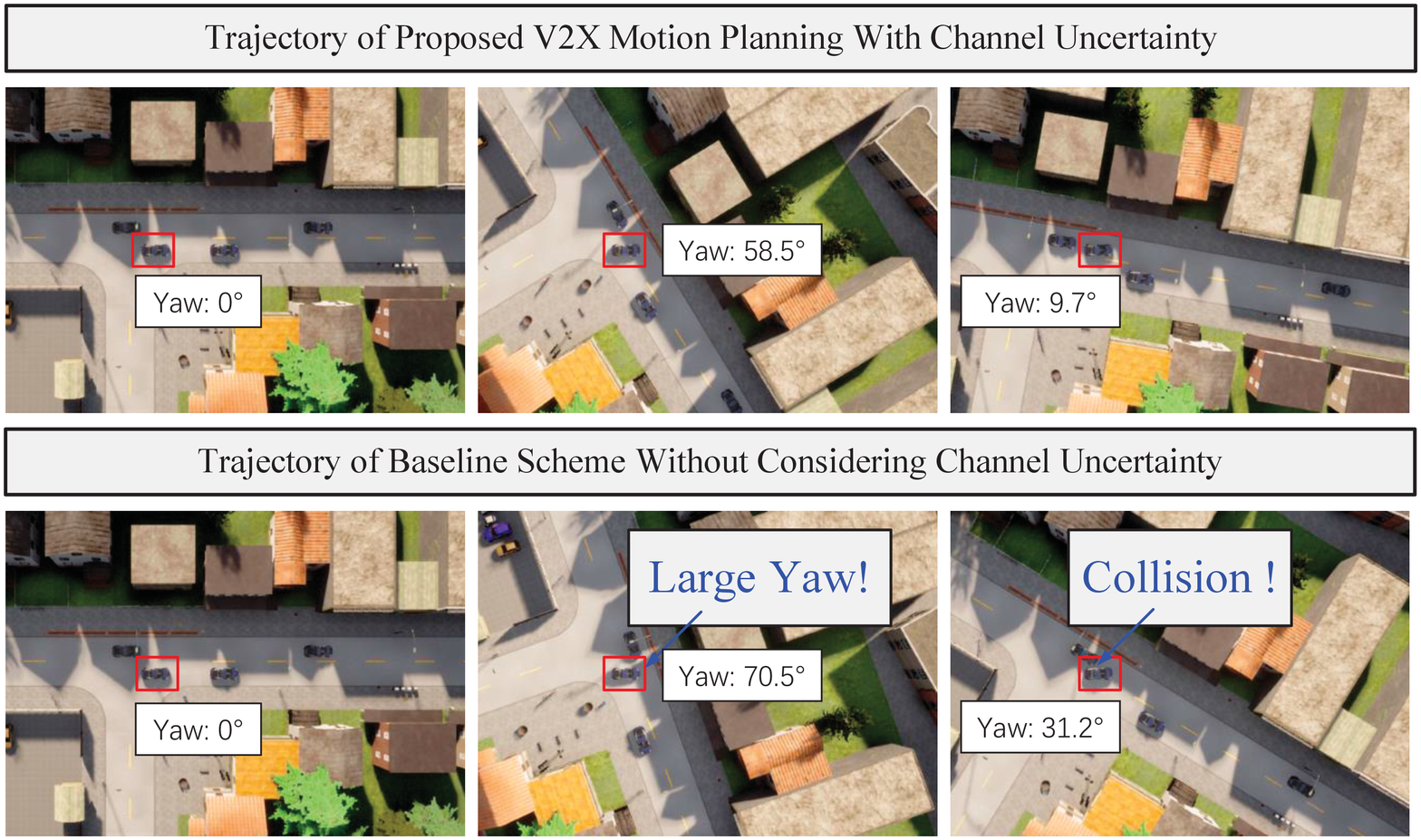}}
			\hspace{0.0in} 
		\subfigure[{Performance comparison versus different outage probabilities with 20 km/h for the LV.}]{ 
		\label{fig:perform_otherBaLiScheme}%% label for first subfigure
		\includegraphics[height=1.2in]{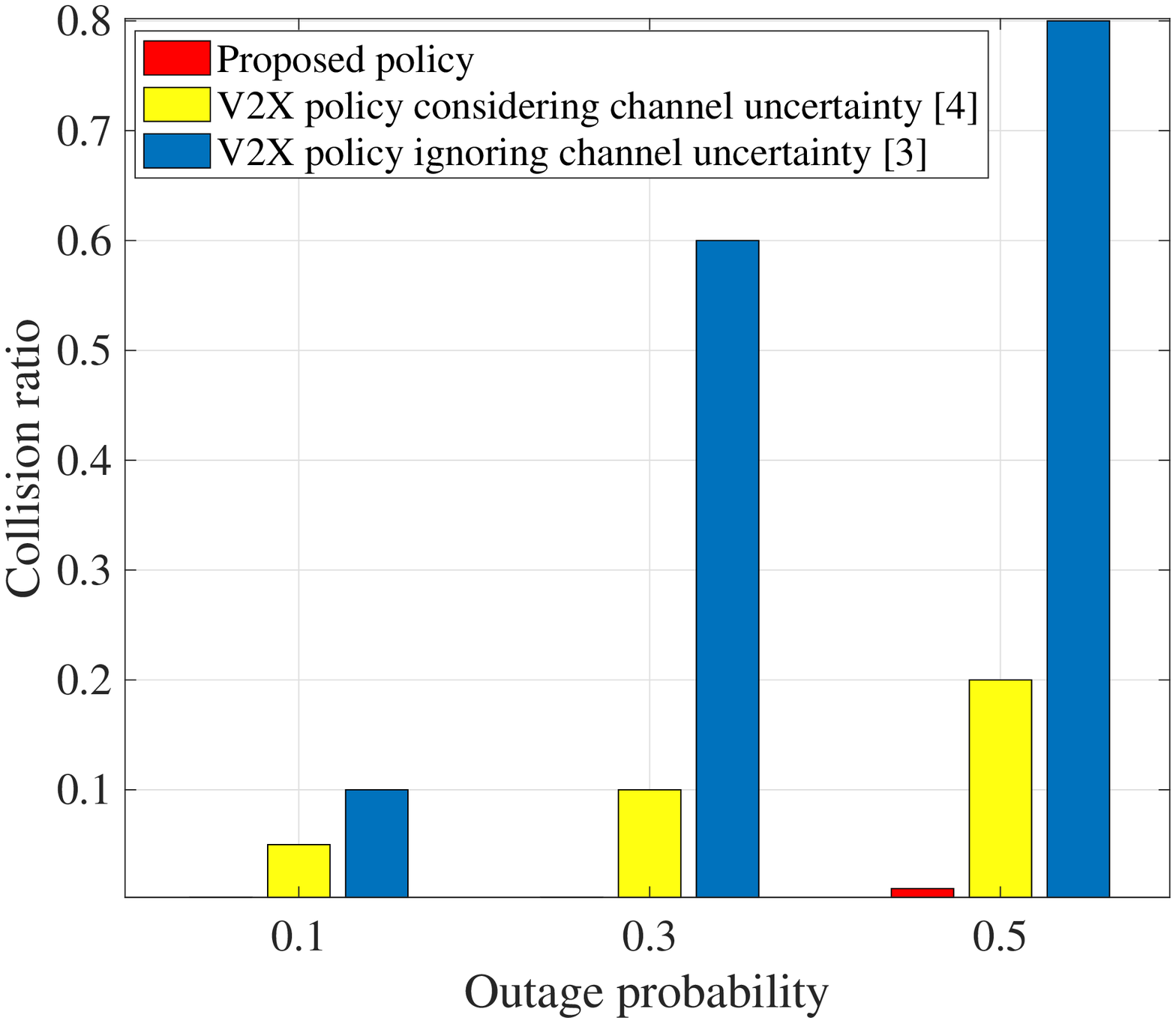}}
	\hspace{0.0in} 
	 \subfigure[{Performance comparison versus different velocities of the LV with 0.3 outage probability}.]{
		\label{fig:Ratio_speed} %% label for second subfigure
		\includegraphics[height=1.2in]{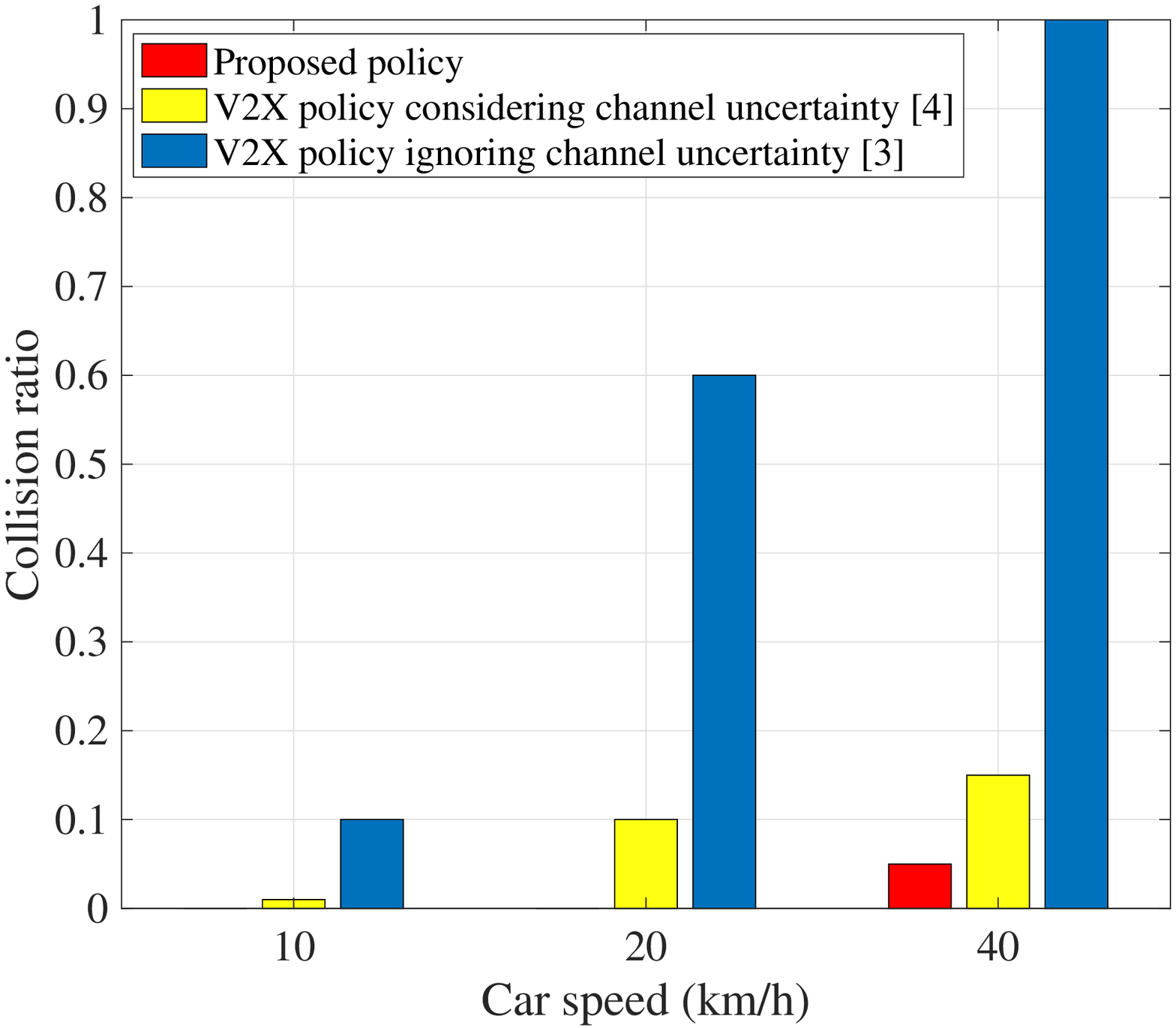}}
 \hspace{0.0in} 
	\caption{Performance comparison of different policies  with $\beta=0.3$ and $a=1$.} 
	\vspace{-0.1in}
\end{figure*}

First, to demonstrate the safety of the proposed policy, we compare its trajectory with that of a baseline policy that ignores the channel uncertainty \cite{baseline1} (i.e., the solution to $\mathcal{P}0$ with $\{\rho_k=0,\forall k\}$). It can be seen from Fig.~\ref{fig:Trajectory_ego}-\ref{fig:Solution_accele_good} that the trajectory of the EV adopting the proposed policy is generally conservative to ensure sufficient distance margin for safety driving compared to that of the baseline policy. As such, the EV successfully bypasses the slow moving LV and avoids collision with the FV during lane-change movement.
In contrast, the baseline policy adopts the competitive driving behavior that leads to a vehicle collision, where the EV collides with the FV at the third time step shown in Fig.~\ref{fig:Trajectory_ego}. Moreover, with higher velocities, the collision time is delayed by a second as shown in Fig.~\ref{fig:Solution_30km}.
It can be seen from Fig.~\ref{fig:Solution_accele_good} that the proposed policy can avoid collision during the lane-change even if the surrounding vehicles are accelerating their speeds.
To clearly explain the implications of relative velocities, the changes of relative velocities are shown in Fig.~\ref{fig:Velocity_Rela}, which indicates that the relative velocities will first decrease (or increase for the TV) and then remain fixed during the turning stage to guarantee a safe following distance.

Next, to account for the vehicle volumes and kinetics in practical environments, the proposed strategy is validated in CARLA simulator \cite{FLCAV}. The experimental results of bird's-eye-views are shown in Fig.~\ref{fig:Collis_outage}, where the EV is marked in a red box. We capture three sequential snapshots of the lane-change procedure with 1 second between consecutive snapshots.
It can be seen from the upper half of Fig.~\ref{fig:Collis_outage} that the EV moves from the ego lane to the target lane, approaching the TV trajectory as time increases, which indicates that the proposed policy plans a dynamically feasible and collision-free trajectory to reach the target lane.
However, the baseline policy, shown in the lower half of Fig.~\ref{fig:Collis_outage}, adopts an exceedingly large yaw angle at the second step, leading to a collision with the FV at the third step. This corroborates the simulation results in Fig.~\ref{fig:Trajectory_ego}, indicating that the proposed policy achieves high safety in both perfect and practical simulators.

Finally, the lane change performance results in terms of the collision ratio during the lane change policy are shown in Fig.~\ref{fig:perform_otherBaLiScheme}-\ref{fig:Ratio_speed}. The collision ratio is defined as the number of collisions divided by the total number of simulation trials during a studied time interval while the simulation results are obtained by averaging over 100 simulation trials. Compared to the V2X policy ignoring the channel uncertainty \cite{baseline1}, the proposed policy significantly reduces the collision ratio, especially under a high outage probability and a high velocity. 
Furthermore, compared with the V2X policy considering the channel uncertainty \cite{baseline2} (to obtain the best performance of \cite{baseline2}, it is assumed that \cite{baseline2} has perfect knowledge of $\tau_{i,k}$), 
the proposed scheme still performs better.
This is because the method in \cite{baseline2} adopts a constant transmit power, while the proposed
scheme allocates more power resources at key waypoints. This finding demonstrates the
necessity of power control under a stringent power budget in V2X motion planning.

\vspace{-0.1in}
\section{Conclusion}
This paper investigated a V2X-aided lane-change scenario in autonomous driving. To ensure safe driving in the existence of channel uncertainties, this paper proposed a robust V2X motion planning strategy by jointly optimizing the safe distance margin, the trajectory, and the transmit power with the BCD approach. Compared to the schemes without 
considering communication delay or adopting power control, the proposed policy significantly reduces the collision ratio in the lane-change scenario.

\vspace{-0.1in}
\bibliographystyle{IEEEtran}

\bibliography{VehicleCom}
\end{document}